\documentclass[runningheads]{llncs}
\usepackage{graphicx}
\usepackage{multirow}
\usepackage[misc]{ifsym}
\begin{document}
\title{MedSyn: LLM-based Synthetic Medical Text Generation Framework}
%
%
\author{Gleb Kumichev\inst{1} \Letter \and
Pavel Blinov\inst{2} \and
Yulia Kuzkina\inst{1} \and 
Vasily Goncharov\inst{1} \and
Galina Zubkova\inst{2} \and
Nikolai Zenovkin\inst{1} \and
Aleksei Goncharov\inst{1} \and
Andrey Savchenko\inst{2}}

\tocauthor{Gleb Kumichev, Pavel Blinov, Yulia Kuzkina, Vasily Goncharov, Galina Zubkova, Nikolai Zenovkin, Aleksei Goncharov, Andrey Savchenko}
\toctitle{MedSyn: LLM-based Synthetic Medical Text Generation Framework}

\authorrunning{G. Kumichev et al.}
%
\institute{MIL Team, Moscow, Russia ~\email{\{gleb.kumichev,yulia.kuzkina,vasily.goncharov,nikolay.zenovkin\}@mil-team.ru} ~\email{alex.goncharov@mil-team.com}\and
Sber AI Lab, Moscow, Russia~\email{\{Blinov.P.D,GVZubkova,AVladSavchenko\}@sber.ru}
}
\maketitle              
\begin{abstract}
Generating synthetic text addresses the challenge of data availability in privacy-sensitive domains such as healthcare. This study explores the applicability of synthetic data in real-world medical settings. We introduce MedSyn, a novel medical text generation framework that integrates large language models with a Medical Knowledge Graph (MKG). We use MKG to sample prior medical information for the prompt and generate synthetic clinical notes with GPT-4 and fine-tuned LLaMA models. We assess the benefit of synthetic data through application in the ICD code prediction task. Our research indicates that synthetic data can increase the classification accuracy of vital and challenging codes by up to 17.8\% compared to settings without synthetic data. Furthermore, to provide new data for further research in the healthcare domain, we present the largest open-source synthetic dataset of clinical notes for the Russian language, comprising over 41k samples covering 219 ICD-10 codes.

\end{abstract}
\section{Introduction}
While extensive open medical datasets are available in English, like the MIMIC family of databases~\cite{mimic4,mimic3} or the CPRD primary care database~\cite{cprd}, their scope in comprehensively covering various medical areas is limited. The availability of textual medical data in non-English languages is even more constrained. Patient privacy and ethical considerations are major limiting factors to the public availability of such data. The latter remains a significant problem; the lack of textual medical resources substantially deters research, testing, and deployment of innovative Natural Language Processing (NLP) methods for national healthcare systems. Synthetic data generation addresses the issue of data scarcity in medical research.

Besides, the population's diseases have a long-tail distribution, with rare diseases representing only a tiny fraction of cases in a dataset~\cite{nguyen2023mimic}. Such data imbalance problem directly affects the ML model's performance on the downstream tasks~\cite{santiso2019class,wang2020utilizing}. Since 2020, our clinical decision support system has been deployed in medical clinics in one of the regions. Insufficient text data on rare cases deters further system scaling, while synthetic (on-demand) medical note generation is the only solution.

\begin{figure}[h!t!]
    \centering
    \includegraphics[scale=0.1]{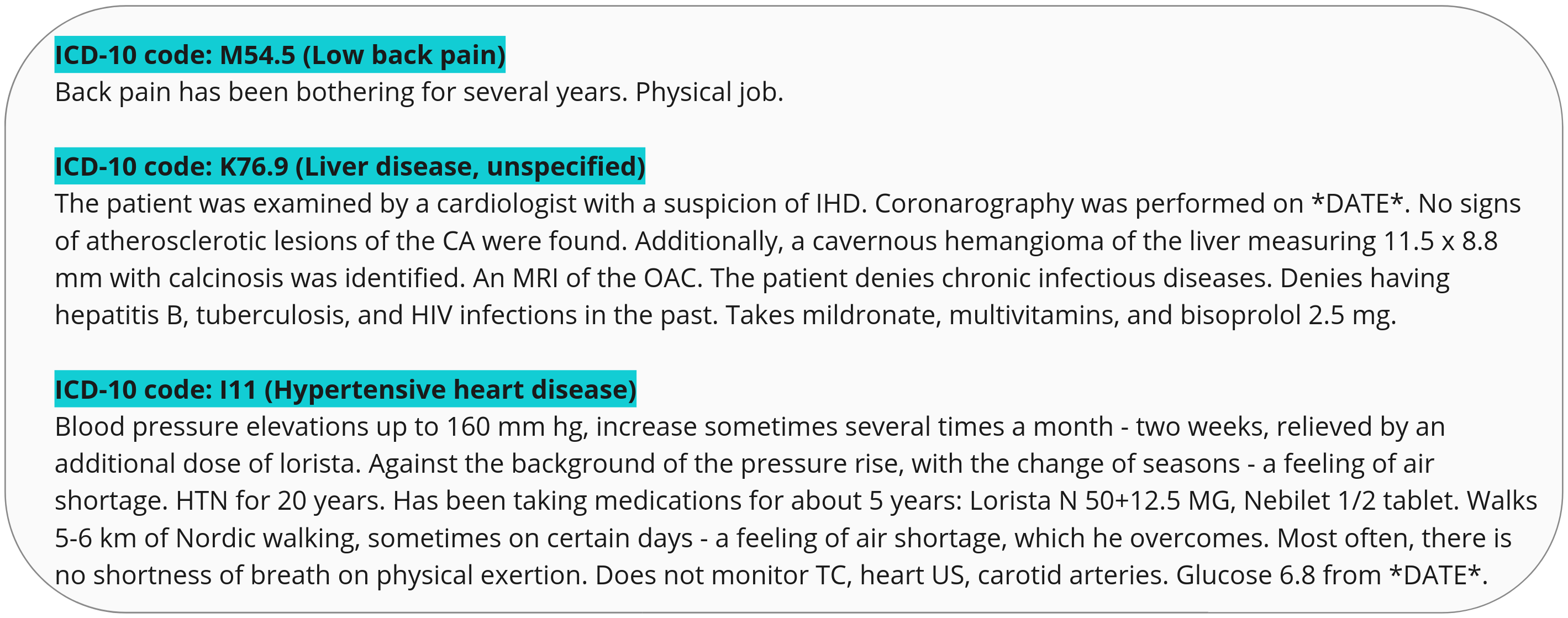}
    \caption{Examples of real clinical notes from RuMedPrime dataset~\cite{rumedprime} (translated to English).}
    \label{fig:hm_example}
\end{figure}

Nowadays, all patient information is stored in Electronic Health Records (EHRs), which contain a structured collection of medical events related to a patient and textual modality attributes: doctor's clinical notes about symptoms and complaints, anamnesis, medication prescriptions, etc. Actual text from clinical notes is a complex object with typos, specialized terms, abbreviations, and contractions. Examples of such notes are shown in Fig.~\ref{fig:hm_example}. That is why some early synthetic generation approaches (e.g.~\cite{pmlr-v68-choi17a}) did not allow dealing with raw clinical text and tried to approximate EHRs only in terms of fixed categorical vectors and a limited set of factors, such as diagnosis and procedure codes or medication names. Including text fragments in synthetic EHRs has been challenging for a long time. Instead of generating medical text from scratch, some proposed frameworks heavily depend on real EHRs~\cite{buczak2010data,reiter2010releasing}, where a new health record is created by data imputation for some critical parts in the original one. However, such an approach limits the variability of results and leaves the risk of private data leakage.

The latest breakthroughs in developing Large Language Models (LLMs) open a new era in generating realistic, coherent, and diverse texts across various domains. Models like GPT-3~\cite{brown2020language}, LLaMA~\cite{touvron2023llama}, and their successors have shown remarkable capabilities in text generation for general and medical texts~\cite{Benoit2023.02.04.23285478}. However, even such powerful models still have some flaws~\cite{ray2023chatgpt}. First of all, they tend to make content errors and hallucinate~\cite{azaria2023internal}, which is unacceptable in such a delicate area as medicine. Therefore, even LLM-based synthetic generation frameworks still need external guidance and internal validation mechanisms to produce medically accurate and relevant texts.

Exploiting Medical Knowledge Graphs (MKGs)~\cite{gao2023leveraging} and ontologies~\cite{Abdollahi2021} is a way to mitigate the problem. Again, such resources are abundant for English but modestly available for less-represented languages like Russian. In this paper, we focus on developing a clinical note text generation framework combining LLMs capabilities with MKG in case studies for the Russian language.

Our key contributions can be summarized as follows:
\begin{enumerate}
    \item We propose an open-source framework called MedSyn\footnote{\label{note1}\url{https://github.com/milteam/MedSyn}} for synthetic clinical note generation. The framework features a novel method that integrates disease-specific symptoms from an MKG and incorporates real data examples into the LLM generation pipeline to enhance the accuracy and diversity of generated data.
    \item We introduce the first dataset\footnote{https://huggingface.co/datasets/Glebkaa/MedSyn-synthetic} with synthetic clinical notes for the Russian language, which contains more than 41k clinical notes spanning over 219 ICD-10 (International Classification of Diseases) codes.
    \item We provide results of experiments on synthetic data generation with the MedSyn framework, including comparisons between GPT-4 and open-sourced LLaMA-7b. It is shown that an open-sourced model fine-tuned on a specific dataset can perform on par with or surpass GPT-4's performance.
\end{enumerate}

\section{Related Work} 
\subsection{Medical Knowledge Graphs} \label{subsection:medical_kg_related}
While a variety of MKGs exist in English~\cite{UMLS,cui2023survey,chandak2023building,10026520}, few or none are available in other languages. There are different possibilities for MGK applications; for example, a line of work utilizes graph embeddings for various medical tasks like recommendation systems~\cite{gong2020smr}, NLI~\cite{Sharma_2019}, and diagnosis prediction~\cite{9679113}. BioLORD~\cite{remy2023biolord2023} uses concepts and relationships from the knowledge graph as part of the LLM pre-training. Another approach for MKG utilization involves enriching the generation process with information extracted from these graphs. This strategy can be viewed as a specialized application of the retrieval-augmented generation framework~\cite{lewis2021retrievalaugmented}, demonstrating the potential to produce more specific, diverse, and factually accurate language. However, applying such techniques in the medical domain is still an area that has not been extensively explored.

\subsection{LLMs in Medical Domain} 
LLMs are increasingly utilized in the medical domain; they are primarily implemented for English~\cite{Luo_2022,singhal2023large,wu2023pmcllama} and Chinese~\cite{zhang2023huatuogpt,xiong2023doctorglm}, evaluated for medical QA tasks, and used as medical chatbots.
There is also a research direction that focuses on synthetic data generation. \cite{peng2023study} trained the GPT-3 model from scratch using clinical and general English texts, then produced 20B of medical texts utilizing this model and introduced a smaller version of the model on synthetic data only. The resulting model outperforms ClinicalBERT~\cite{huang2020clinicalbert} and the same model trained on actual data on MedNLI~\cite{romanov2018lessons} and emrQA~\cite{pampari2018emrqa} benchmarks.
The authors of~\cite{article} generated clinical texts and manually annotated them for the Named Entity Recognition (NER) task. The evaluation shows that the combination of original and synthetic corpora achieved better performance than using only the initial corpus. In~\cite{tang2023does}, the authors improve performance on NER and relation extraction tasks with synthetic data, showing that increasing the number of synthetic sentences can improve model performance up to a certain point, beyond which the improvement becomes marginal. In a recent study~\cite{hiebel-etal-2023-synthetic}, researchers explored the feasibility of using synthetic text as a training corpus for clinical NER in French. The findings suggest that synthetic clinical notes can be used to train NER models, although applications for other tasks remain to be explored.

The true potential of synthetic data in the medical field remains under active exploration~\cite{shaib2023summarizing,tang2023does}. However, typical problems related to LLMs, like hallucinations, pose substantial challenges in such a critical field. Ensuring factual accuracy and addressing inconsistencies in medical models remain valuable concerns~\cite{Xie2023.04.18.23288752}. In our research, we strive to bridge the gap in controllable medical data generation, primarily focusing on the Russian language, which is heavily underrepresented in linguistic medical resources.

\begin{figure*}[t]
    \centering
    \includegraphics[width=\textwidth]{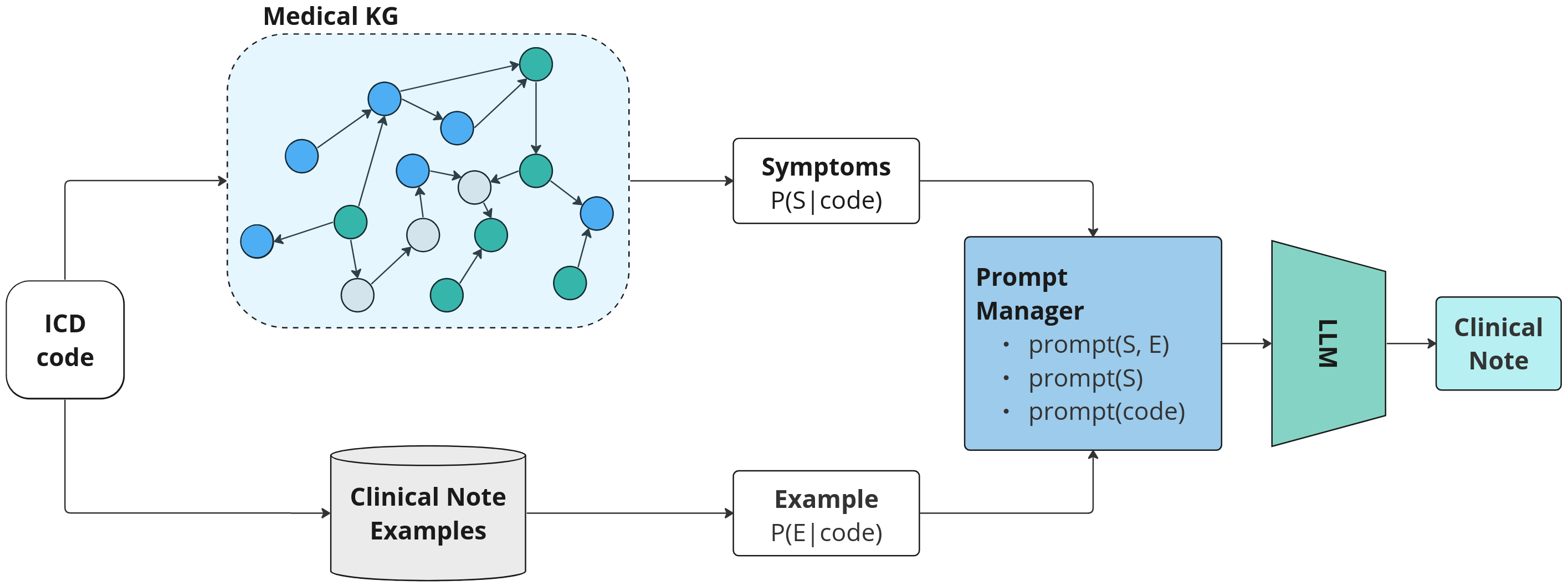}
    \caption{The clinical notes generation pipeline implemented in MedSyn framework. Relevant symptoms from MKG and clinical note examples corresponding to the ICD code are compiled into a prompt and used as input for LLM inference.}
    \label{fig:pipeline}
\end{figure*}

\section{Method}
The overall pipeline for clinical note generation is illustrated in Fig.~\ref{fig:pipeline}. To generate a clinical note for a target ICD code, data relevant to the MKG (Section~\ref{subsection:medical_kg}) and real examples are first sampled and combined into a prompt for LLM inference. We utilized GPT-4 and a fine-tuned LLaMA-7b for the LLMs (Section~\ref{subsection:models}). For fine-tuning LLaMA-7b, we constructed an instruction-following dataset (Section~\ref{subsection:ft_dataset}). To generate a dataset of clinical notes for our experiments, we developed a specific generation task (Section~\ref{subsection:gen_task}) with already prepared prompts.

\subsection{Medical Knowledge Graph} \label{subsection:medical_kg}
As mentioned in Section~\ref{subsection:medical_kg_related}, Russian-language equivalents of MKG are scarce. For our research, we used the WikiMed database as a foundation to develop the Russian MKG.

\begin{table}[htb]
\centering
\caption{MKG statistics. Di-Dr stands for disease-drug relation, Di-S for disease-symptom relation.}
\begin{tabular}{cccc|cc}
    \hline
    & \multicolumn{3}{c|}{\textbf{Nodes}} & \multicolumn{2}{c}{\textbf{Edges}} \\
    & \textit{Disease} & \textit{Drug} & \textit{Symptom} & \textit{Di-Dr} & \textit{Di-S} \\
    \hline
    \textbf{\#} & 2,747 & 2,968 & 2,554 & 1,997 & 2,554 \\
    \hline
\end{tabular}
\label{tab:kg_stats}
\end{table}

The constructed MKG includes the following nodes: diseases (identified by ICD-10 codes), drugs, and symptoms. While diseases and drugs have predefined relations in this database, symptoms and their relations are not specified. The database includes clinical manifestations, which contain potential symptoms in a narrative format. To extract these symptoms, we utilized ChatGPT~\cite{openai2023gpt4}, prompting it to identify symptoms from the given text of clinical manifestations.
For example, the clinical manifestation of tuberculosis, \textit{`One of the common manifestations of spinal tuberculosis is the formation of cold abscesses on the neck and increased skin temperature`}, should lead to the extraction of symptoms \textit{[cold abscesses on the neck, increased skin temperature]}.
The extracted data were manually verified by comparing them with the initial text to ensure that only symptoms were included, and no irrelevant information or noise was extracted.

Finally extracted symptoms were then incorporated into the MKG. Its statistical details are presented in Table~\ref{tab:kg_stats}.

\begin{figure}
    \centering
    \includegraphics[scale=0.08]{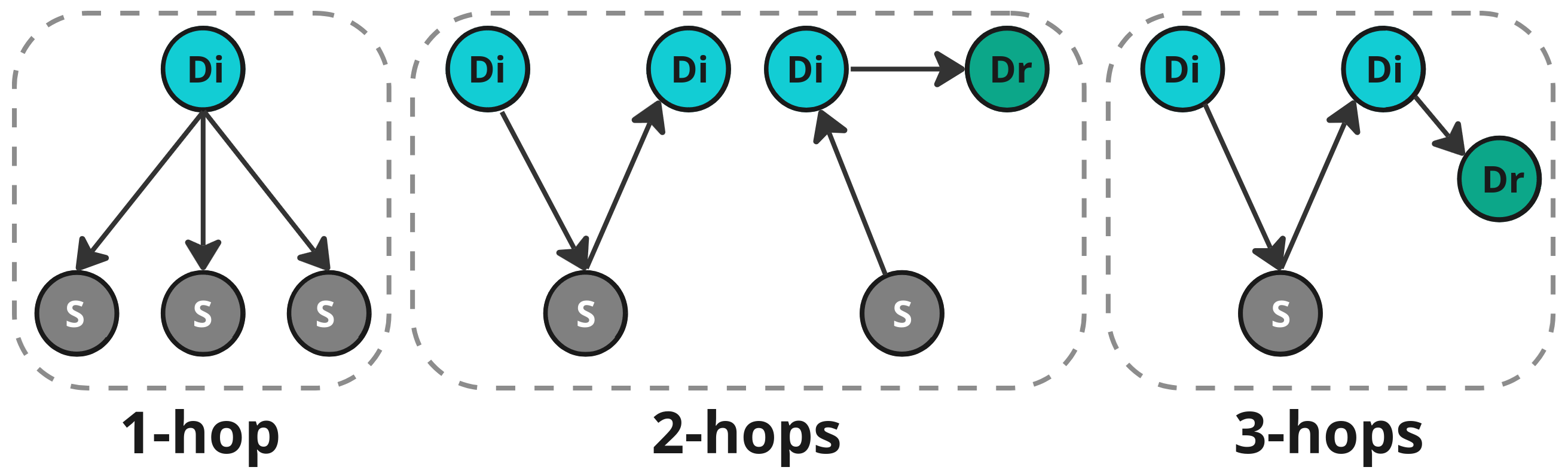}
    \caption{Examples of k-hop reasoning question on MKG. Di - Disease, Dr - Drug, S - Symptoms.}
    \label{fig:hop_q}
\end{figure}

\subsection{Instruction-Following Dataset} \label{subsection:ft_dataset}
We collected a dataset of 152k Russian language samples focused on instruction-following for supervised fine-tuning\footnote{https://huggingface.co/datasets/Glebkaa/MedSyn-ift}. These samples were derived from various medical benchmarks, databases, and the constructed MKG. Utilizing the MKG, we created questions that require multiple levels of reasoning, ranging from simple 1-hop to complex 3-hop distances. For example, a 1-hop reasoning question like \textit{'Provide symptoms for a disease'} directly connects diseases to symptoms (Di-S). A 2-hop question, such as \textit{'Write down medications that can be taken for these symptoms'}, involves linking symptoms to diseases and then to drugs (S-Di-Dr). A more complex 3-hop reasoning question, like \textit{'List medications that can be taken for a disease if it is mistaken for another disease with similar symptoms'}, maps diseases to symptoms, then to another disease, and finally to drugs (Di-S-Di-Dr), as shown in Fig.~\ref{fig:hop_q}. 
We avoid more than three hops reasoning scenarios as, by our estimate, it produces too vague and error-prone samples. For the clinical notes, we employed two types of tasks: continuation, which extends an existing note from a random point, and generation, where a note is created from prior data like symptoms. We generated at least five different rephrasings for each to ensure instruction diversity.

\begin{figure}
    \centering
    \includegraphics[scale=0.12]{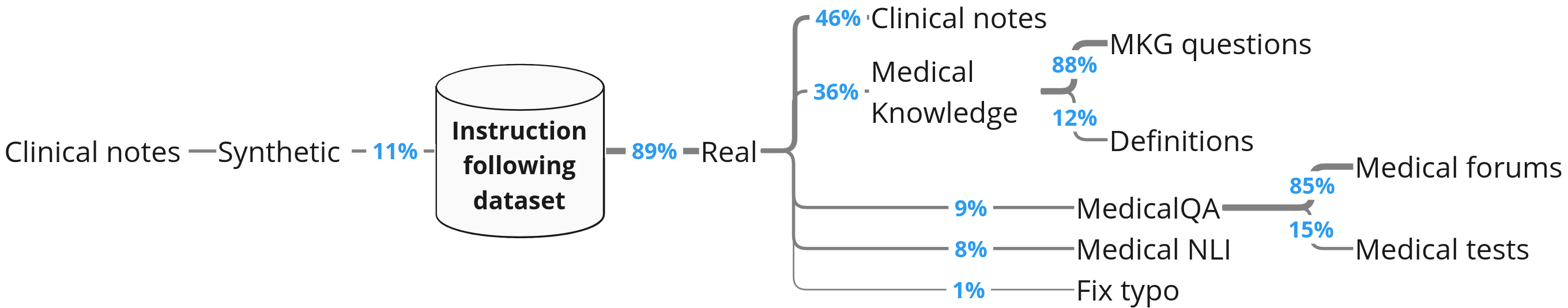}
    \caption{The structure of the instruction-following dataset. Leaves represent data sources and the percentage of data relative to the parent category.}
    \label{fig:data-stats}
\end{figure}

In addition to real medical data, we also incorporate synthetic data from ChatGPT. Considering that real clinical notes often have many typos and stylistic variations, which may affect model performance, we suggest that adding synthetic notes could improve the model's text generation and be a regularization method. To create this synthetic data, we prompted ChatGPT to generate clinical notes based on patient symptoms, age, and gender. For part of the data, style references with real samples were additionally provided. We also incorporate a medical dataset focused on typo correction to make the model more robust to typos. The structure of the dataset is represented in Fig.~\ref{fig:data-stats}. 

\subsection{Fine-Tuning} 
\label{subsection:models}
Unlike the English language, to our knowledge, there aren't any open-source generative LLMs tailored for the medical domain in Russian. Thus, we employ GPT-4~\cite{openai2023gpt4} for data generation to establish a strong baseline. 

Our work uses a model based on the LLaMA 2 family~\cite{touvron2023llama}. It is a collection of open generative language models with a parameter range from 7 to 70 billion. We fine-tuned the model with 7 billion parameters using a learning rate $2e^{-5}$ and a cosine learning rate scheduler to fine-tune the model. We utilized a global batch size of 256 and trained the model for three epochs.

To enhance the efficiency and accelerate the training of our model, we employed Low-Rank Adaptation (LoRA)~\cite{hu2021lora}. This method involves freezing the model's weights and injecting trainable rank decomposition matrices into each layer of the Transformer architecture.

The pre-training data for LLaMa-7b consists of 90\% English-language data and only 0.13\% Russian-language data. Therefore, to fine-tune our model, we decided to use the pre-trained checkpoint from Saiga 2\footnote{\url{https://huggingface.co/IlyaGusev/saiga2\_7b\_lora}} that is fine-tuned on Russian language instructions and dialogues generated by GPT-4.

\subsection{Generation Task} \label{subsection:gen_task}
We prepared a generation task to generate synthetic clinical notes with real data examples and symptoms spanning 105 ICD-10 category codes, as presented in the RuMedTop3 dataset~\cite{Blinov_2022}.
We sample symptoms previously extracted from Russian MKG (Section~\ref{subsection:medical_kg}) according to the approach outlined in Section~\ref{subsection:Symptoms_Sampling}.

We aim to achieve a uniform distribution of ICD codes for the generation task, but the lack of data requires inevitable trade-offs. Given the limited set of examples (1,283 samples), and to ensure that the sampling procedure represents the diversity of examples and symptoms, we have adopted a specific approach to determine the frequency of each ICD-10 category code $\mathcal{C}$ and computes its weight based on the following rule:

\begin{equation}\label{eq:eq_2}
    w(\mathcal{C}) =
    J_3\!\left(N^{\mathcal{C}}_{\mathrm{symp}}\right) \cdot J_3\!\left(N^{\mathcal{C}}_{\mathrm{exmp}}\right),
\end{equation}
where $J_3$ denotes the triple application of the function $J(x) = \log(1 + x)$, $N^{\mathcal{C}}_{\mathrm{exmp}}$ refers to the number of examples corresponding to a given category code $\mathcal{C}$, and $N^{\mathcal{C}}_{\mathrm{symp}}$ represents the number of all unique symptoms within that category. The logarithmic scale used in Eq.~\ref{eq:eq_2} is implemented to achieve a more uniform distribution of codes.

An exception to this weighting procedure is the category \textsc{Z00}, defined as \emph{encounter for the general exam without complaint, suspect, or reported diagnosis}. As this category does not hold particular interest for downstream tasks, we set the number of generations for this category code to 10, thereby not factoring its weight into the overall distribution. We obtained the final generation task by sampling clinical notes and symptoms for this distribution, containing 2,503 entries. Each entry consists of an ICD-10 code, an example of a real clinical note, and a subset of symptoms.

For the baseline, we generate samples that do not utilize data from MKG in their prompts. The baseline prompt is similar to the original one but contains only the disease name instead of incorporating disease prior information from MKG and a clinical note example. Generated and real clinical notes contain no ICD codes in the text to avoid data leaks.

\subsection{Symptoms Sampling} \label{subsection:Symptoms_Sampling}
The actual distribution of symptoms in clinical settings is complex. For example, certain symptoms may not coexist or be specific to a particular age or gender. In this study, however, we assume that symptoms are independently and identically distributed. Consequently, we select multiple symptoms for a disease without considering their inter-relationships. We randomly sample several symptoms from the MKG (Section~\ref{subsection:medical_kg}) related to a disease, with the count ranging from 1 to 5, which is also chosen randomly.

\subsection{Synthetic Dataset} \label{section:synt_dataset}
We have released a dataset of 41,185 synthetic clinical notes in Russian, generated using GPT and fine-tuned LLaMA models spanning 219 ICD-10 codes. The dataset includes all generated samples, regardless of quality, to facilitate various data selection methods. More detailed statistics and descriptions of the data fields are provided in the project dataset repository.\footnote{\label{note2}\url{https://huggingface.co/datasets/Glebkaa/MedSyn-synthetic}} According to the provided licenses, all confidential information was anonymized, and researchers can safely use these datasets.

\section{Experiments}
\subsection{Datasets and Tasks}
In this research, we utilized the RuMedPrime dataset~\cite{rumedprime}, containing 7,625 anonymized entries from outpatient visits to the Siberian State Medical University hospital. This dataset, unique as the only open-source collection of clinical notes in Russian annotated with ICD-10 codes, comprises each patient's clinical note, symptoms, and corresponding ICD code. Based on this dataset the RuMedTop3 task was created, focusing on the ICD code prediction from a free-text clinical note. Given such a task, it is possible to implement an AI service that supports doctors with a second opinion on the diagnosis search.

Our study adopted the same dataset split as RuMedTop3, using 4,690 records for training, 848 for validation, and 822 for testing while incorporating full clinical notes alongside symptoms. Like RuMedTop3, we employ the second ICD-10 classification code hierarchy level. We also evaluated the results on the original RuMedTop3 dataset.

\subsection{Models}
We conducted experiments using both feature-based linear models and transformer models. For the linear model, we employed logistic regression based on term frequency-inverse document frequency (TF-IDF) features. For the transformer models, we run experiments with RuBERT~\cite{kuratov2019adaptation} and RuBioRoBERTa~\cite{yalunin2022rubioroberta} and report the average results from three runs.

\subsection{Evaluation}
ICD code prediction is a multi-class classification task. To evaluate it, we utilize the $hit@k$ score ($k \in [1, 3, 5]$), defined as follows:
\begin{equation}\label{eq:topk}
    hit@k = \frac{1}{N} \sum_{i=1}^{N} hit(\hat{y}, top_i^k),
\end{equation}
where $N$ is the number of samples and $hit(\hat{y}, top_i^k)$ is 1 if ground truth ICD code $\hat{y}$
is on a ranked list of $k$ predicted codes $top^k$ and 0 otherwise.

\subsection{Results}
\subsubsection{Prompt Following}

\begin{figure}[htp]
    \centering
    \includegraphics[scale=0.45]{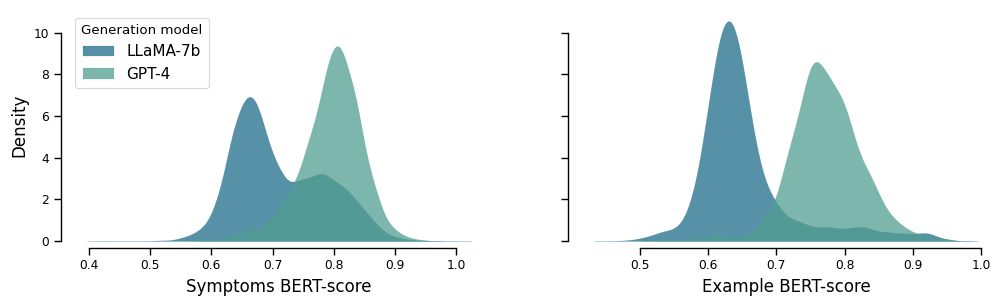}
    \caption{BERT-scores for example and symptoms usage.}
    \label{fig:bert-scores}
\end{figure}

We use the BERT-score~\cite{zhang2020bertscore} to measure the similarity of synthetic data to the examples and to the provided symptoms (Fig.~\ref{fig:bert-scores}).

As can be seen from the higher scores, the GPT-4 model follows instructions more precisely, produces results that are more similar to the example, and makes greater use of the provided symptoms.

\begin{figure}[htp]
    \centering
    \includegraphics[scale=0.45]{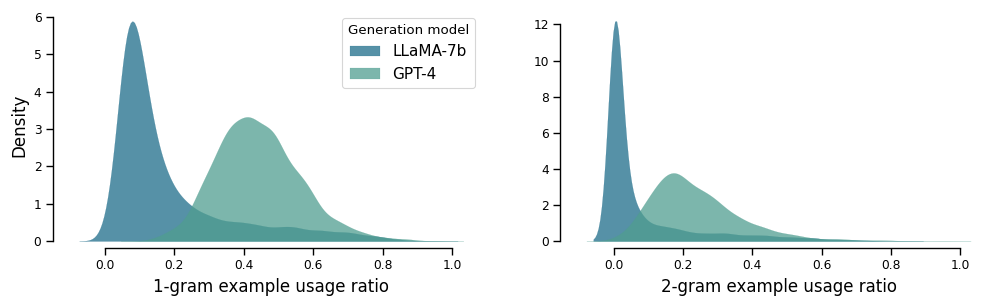}
    \caption{Ratio of N-gram usage.}
    \label{fig:example-usage-n-gram}
\end{figure}

While high similarity to the example is desirable, complete replication is unfavorable. To evaluate replication, we calculate the ratio of example N-grams usage, defined as the ratio of unique common N-grams between the generated sample and the example, divided by the number of unique N-grams in the example (Fig.~\ref{fig:example-usage-n-gram}). For most samples, the N-grams usage ratio is less than 1, suggesting that the examples are far from complete replication in the answer.

\subsubsection{Generating Data Out of the Training Set} \label{subsection:oots-data}

\begin{figure}[h!tb]
    \centering
    \includegraphics[scale=0.45]{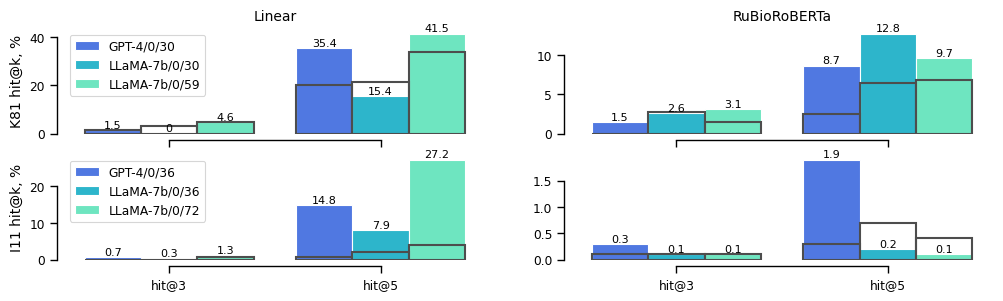}
    \caption{The prediction results using only synthetic training data (codes K81 and I11). Contour bars represent the baseline prompt, which does not utilize MKG and consists solely of the task and the disease name.}
    \label{fig:code_replacement-K81-I11}
\end{figure}

One of the most exciting yet practically challenging scenarios involves generating data scarcely present in the original training set or generation of clinically valuable data. We selected two vital ICD codes for the experiment, K81 and I11. The first is cholecystitis, which affects about 20\% of the adult population. The second code denotes a type of heart disease, one of the most common causes of death.

We transferred all real data samples to the test set, making evaluating the experiments with real data in the training set impossible. However, we prioritize a diverse test set in this experiment as it could mitigate the potential poor performance of unrepresented synthetic samples in downstream tasks.
We replaced the real data in the training set with 30 synthetic samples for both models and added 59 samples for LLaMA-7b to assess the impact of scaling the number of samples (Fig.~\ref{fig:code_replacement-K81-I11}).

Although models trained with such synthetic data still have zero scores in the $hit@1$ metric, they show promising results in less restricted metrics like $hit@5$, demonstrating the potential for further improvement in real data absence scenarios. Thus, synthetic data with specific refinements could increasingly become a viable alternative for training models in data-scarce environments.

\subsubsection{Synthetic Upsampling}
\begin{figure}[h!tb]
    \centering
    \includegraphics[scale=0.45]{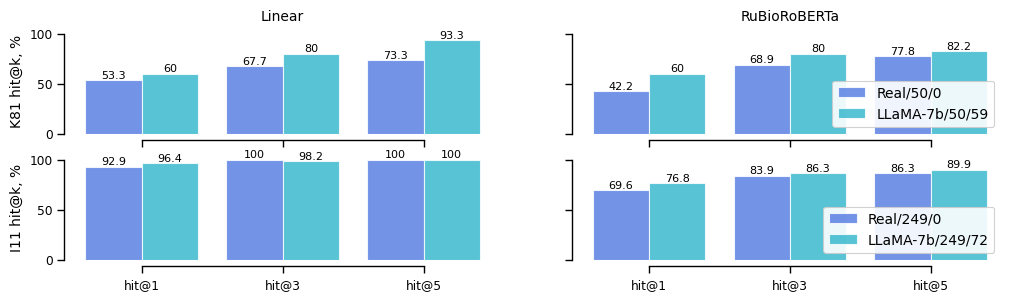}
    \caption{Results of prediction with upsampled training set for codes K81 and I11. The legend represents the data source/number of real samples/number of synthetic samples.}
    \label{fig:code_upsampling-K81-I11}
\end{figure}

Another application for synthetic data is data upsampling. In this experiment, we used the same synthetic data as in the previous section (Section~\ref{subsection:oots-data}) and added it to the training set. The results indicate that models can benefit from such synthetic data. For instance, the accuracy of K81 code prediction improves by 17.8\% for the RuBioRoBERTa model (Fig.~\ref{fig:code_upsampling-K81-I11}). To assess the overall accuracy across all ICD codes, we also evaluate both the baseline and the full prompts (Table~\ref{tab:rumedtop3-ups-table}).

\begin{table}[h!tb]
\centering
\caption{Scores across all codes with upsampled training set for codes K81 and I11.}
\begin{tabular}{{l}{l}{c}{c}{c}{c}}
\hline
\textbf{Model} & \textbf{Data} & \textbf{hit@1} & \textbf{hit@3} & \textbf{hit@5} \\
\hline
\multirow{3}{*}{Linear} & Real  &56.9 &\textbf{80.4} &87.5\\
                        &LLaMA-7b-baseline  &57.1 &80.0 &\textbf{93.9} \\
                        &LLaMA-7b  &\textbf{57.2} &79.9 &93.7 \\
                        \hline
\multirow{3}{*}{RuBioRoBERTa} & Real &52.7 &75.8 &84.3 \\
                        &LLaMA-7b-baseline  &55.8 &75.1 &82.8 \\
                        &LLaMA-7b  &\textbf{56.8} &\textbf{77.7} &\textbf{86.2} \\
                        \hline
\end{tabular}
\label{tab:rumedtop3-ups-table}
\end{table}

For a more detailed analysis, we focused on two codes that were frequently mistaken for each other more than any other pair. This decision was based on the confusion matrix, which measures how often each pair of codes is confused. The analysis revealed that the codes most often confused are M54 and G54.

We selected synthetic data for those codes generated via the same generation task described in Section~\ref{subsection:gen_task} for the GPT-4 and LLaMA-7b models. For the LLaMA, we repeated the generation several times to evaluate the effect of data scaling. Here, we only report on the linear model to depict simultaneous changes for codes not averaged across several models. The experimental results are presented in Table~\ref{tab:m-code-result}. While data generated by GPT-4 provides improvements for both codes simultaneously, data generated by LLaMA still offers improvement for one of the codes without a drop for the other.

\begin{table}[htb]
\centering
\caption{Results of upsampling for the most pairwise misclassified codes. Prediction by the linear model. \#R/S represents the number of real and synthetic samples in the training set. $\uparrow$ represents growth of both codes simultaneously, $\nearrow$ - growth for one code without drop for another.}
\begin{tabular}{{c}{l}{c}{c}{c}{c}}
\hline
\textbf{Code} & \textbf{Data} & \textbf{\# Real/Synthetic} & \textbf{hit@1} & \textbf{hit@3}  & \textbf{hit@5} \\
\hline
\multirow{4}{*}{G54} &  Real & 232/0 &57.9	&97.4  &100\\
                               &GPT-4 & 232/14 &60.5 	$\uparrow$  &97.4 &100 \\
                               &LLaMA-7b & 232/14 &57.9 &97.4 &100 \\
                               &LLaMA-7b & 232/72 &60.5 $\nearrow$ &97.4 &100 \\
                                \cline{2-6}
\multirow{4}{*}{M54} & Real & 560/0 &85.1	&98.9 &100 \\
                          &GPT-4 & 560/35  &87.4 	$\uparrow$ &98.9 &100 \\
                          &LLaMA-7b & 560/35 &86.2 $\nearrow$ &98.9 &100 \\
                          &LLaMA-7b & 560/175  &85.1	&98.9 &100\\
                        \hline
\end{tabular}
\label{tab:m-code-result}
\end{table}

\subsubsection{RuMedTop3 Upsampling}
Although the generated clinical notes contain more information than the data in the RuMedTop3 task, which focuses on symptoms, using the generated data to upsample this dataset is still feasible, as they share the same set of ICD codes. We report results with generated data upsampling in Table~\ref{tab:rumedtop3-ups}, showing that all models benefit from the synthetic data. 

\begin{table}[h!tb]
\centering
\caption{Results of upsampling on RuMedTop3 dataset (the real data size is 4,690 samples, and the size of the synthetic dataset is 2,503 samples).}
\begin{tabular}{{l}{l}{c}{c}{c}{c}}
\hline
\textbf{Model} & \textbf{Data} & \textbf{hit@1} & \textbf{hit@3} & \textbf{hit@5} \\
\hline
\multirow{3}{*}{Linear} & Real  &49.8 &72.7 &87.8	\\
                        &GPT-4  &\textbf{50.8} &\textbf{74.8} &\textbf{90.0} \\
                        &LLaMA-7b  &50.2 &73.6 &89.5 \\
                        \hline
\multirow{3}{*}{RuBERT} & Real &46.5 &70.4 &79.3 \\
                        &GPT-4  &\textbf{47.2}	 &\textbf{71.9} &81.3 \\
                        &LLaMA-7b &45.0 &70.7 &\textbf{81.4} \\
                        \hline
\multirow{3}{*}{RuBioRoBERTa} & Real &\textbf{47.4} &70.8 &79.5	\\
                        &GPT-4  &47.3	 &\textbf{71.7} &\textbf{80.4} \\
                        &LLaMA-7b  &46.1 &70.4 &79.6 \\
                        \hline
\end{tabular}
\label{tab:rumedtop3-ups}
\end{table}

\subsection{Human assessment}
We performed the human evaluation in a side-by-side scenario to qualitatively assess the synthetic clinical texts. First, we randomly sampled 105 cases from real clinical notes examples according to the general ICD code distribution and paired them with synthetic ones. Second, in each pair, we selected random sentences (with a median number of words of 8) to facilitate labeling and make a fair comparison detached from the notes structure. Such text pairs were presented to a medical intern with the only question -- \textit{Which text is generated, 1 or 2?} The assessor was correct in 58.09\% (61 cases). Given that the random guessing is 50\%, we can conclude that our synthetic texts have acceptable quality. In further research, we plan to evaluate the MedSyn framework in more elaborate human assessment scenarios.

\section{Discussion}
We used the generated datasets during all evaluations without applying filtration or sample selection techniques. Consequently, these datasets likely contain corrupted samples with minor factual errors or in some kind irrelevant to the provided prompt. 

To estimate the validity of the samples, we predict their label using models trained on real data. We calculate the ratio of valid samples whose ground truth label appears in the top 5 predictions of at least 2 of five RuBERT models, each trained with different seeds. We found that 51\% of LLaMA-7b samples and 64\% of GPT-4 pass this criterion. However, this is only a coarse criterion as it may lead to false negatives, where a correct synthetic sample falls outside the training distribution and is consistently misclassified. Additionally, a sample might contain relevant information that leads to accurate predictions while still having some corruption. This observation also suggests that GPT-4 generated data might include fewer inaccurate samples, contributing to better performance. Possible sample corruption could lead to gaps in the authenticity and applicability of the generated content in specific clinical scenarios, highlighting the need for advanced filtration algorithms to refine the data quality. Future enhancements to the MKG, including a broader range of medical information, will likely improve the robustness and diversity of generated synthetic data.

While synthetic data is not directly tied to real patients, its use in clinical settings can still pose ethical questions regarding its applicability and acceptability. Key concerns include: 1) Ensuring that the data accurately reflects diverse patient populations without introducing biases; 2) Protecting against potential indirect privacy violations; 3) Assessing how its use might impact clinical decision-making.
Additionally, it is essential to be transparent about how synthetic data is made and used and ensure its use follows informed consent rules in medical settings.

\section{Conclusion}
The proposed MedSyn framework suggests promising results in generating synthetic clinical notes. Human evaluation shows the high quality of generated texts, which are indistinguishable from real medical notes. In numerical experiments, using additional synthetic notes leads to a 17.8\% increase in ICD-code classification accuracy for vital and challenging classes compared to using only real data. Additionally, models trained on generated data reveal substantial quality even when used as the only training source, beating a solid baseline and helping to improve scores on the RuMedTop3 task. From a practical point of view, we plan to exploit the developed framework for rare disease note generation. Such synthetic data will allow us to substantially increase the number of disease classes in our clinical decision support system from tens to hundreds of ICD codes, giving the doctor a reliable second opinion even in rare scenarios.

The framework's design allows easy integration with diverse MKGs, promising even more robust and varied data generation. To foster continued innovation in this field, we have made our trained model, part of the training dataset, and the synthetic dataset publicly available. These resources pave the way for further research in the medical field, especially in tasks where data is scarce. For instance, they potentially serve as datasets for medical NER tagging or in ICD coding tasks, where models trained on such data could provide valuable automated suggestions to humans. While synthetic data may contain inconsistencies or flaws, it is still precious in low-resource languages (like Russian) or low-data areas (like healthcare).

%
%

\end{document}